\documentclass[11pt]{article}

\usepackage[final]{acl}

\usepackage{times}
\usepackage{latexsym}

\usepackage[T1]{fontenc}

\usepackage[utf8]{inputenc}

\usepackage{microtype}

\usepackage{inconsolata}
\usepackage{colortbl}
\usepackage{xcolor}
\usepackage{graphicx}
\usepackage{textcomp}
\usepackage{xcolor}
\usepackage{booktabs}
\usepackage{multirow}  
\usepackage{siunitx}  
\usepackage{subcaption} 
\usepackage{tabularx}
\usepackage{makecell}

\title{MAVIS: Multi-Agent Video Retrieval via Structured Video Understanding}

\author{Jie Zhang$^1$ , Qilang Ye$^{2}$, Hao Zhou$^{1,3}$, Haochen Liang$^4$, Fei Luo$^{1}$\thanks{Corresponding Author}\\
         $^{1}$School of Computing and Information Technology, Great Bay University  \\ $^{2}$College of Computer Science, Nankai University\\ 
         $^3$Tsinghua Shenzhen International Graduate School, Tsinghua University \\
         $^4$Graduate School of Information Science and Technology, The University of Tokyo \\ jz@stu.cqut.edu.cn, luofei@gbu.edu.cn}

\begin{document}
\maketitle
\begin{abstract}
The dominant paradigm in video retrieval relies on embedding-based full-corpus scanning, which suffers from inherent computational inefficiency and the semantic asymmetry between information-dense videos and sparse textual queries. To bridge this gap, we introduce \textbf{MAVIS}, a novel multi-agent framework that rethinks retrieval as cooperative reasoning rather than brute-force search. MAVIS first bridges the granularity mismatch by parsing raw videos into a \textbf{Structured Semantic Library}, enabling explicit attribute-level indexing. During retrieval, a planner decomposes complex user intents into atomic sub-tasks, dispatching specialized agents to independently nominate candidates. Crucially, MAVIS employs a \textbf{Logic-aware Debate} mechanism with a strict veto protocol, where agents collaboratively prune logical mismatches to identify a compact set of ``controversial'' candidates for fine-grained verification. This agentic workflow effectively bypasses the inefficiency of full-library traversal. Extensive experiments on MSR-VTT, MSVD, and ActivityNet demonstrate that MAVIS achieves competitive performance without task-specific fine-tuning, offering a scalable and interpretable alternative to traditional dual-encoder approaches.
\end{abstract}

\section{Introduction}
The exponential growth of video content on the web has precipitated an urgent demand for intelligent retrieval systems capable of navigating massive multimedia archives~\cite{wang2025koala}. While recent years have witnessed significant strides in video retrieval, the dominant paradigm remains \textit{embedding-based full-matching}, where every query is compared against the holistic feature representations of all videos in the corpus~\cite{wang2024text, tian2024holistic, tang2025muse}. This brute-force approach encounters a fundamental bottleneck: \textbf{semantic asymmetry and computational redundancy}. User queries are typically compact and intent-specific (e.g., ``a red car turning left''), whereas videos are information-dense streams containing complex, multifaceted semantics. 
Relying solely on global embeddings for retrieval across the entire corpus is inherently prone to noise~\cite{ma2022x, tian2024holistic}; 
the subtle visual cues required by the query are often statistically drowned out by the vast number of irrelevant distractors and the dominant background features of the video itself.
Moreover, exhaustively scanning the entire library for every query incurs prohibitive computational costs. Consequently, there is a pressing need for a paradigm shift from ``brute-force scanning'' to ``structured reasoning.''

\begin{figure}[t]                    
  \centering
  \includegraphics[width=1.0\columnwidth]{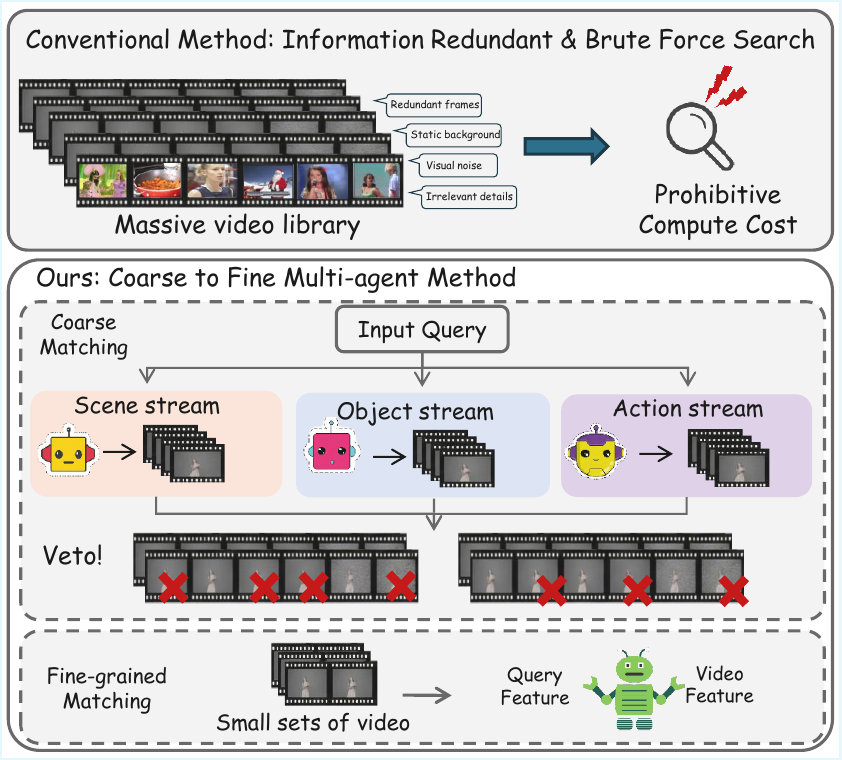}
  \caption{Overview of MAVIS. Specialized agents tackle distinct dimensions to prune the search space through independent proposal and collaborative debate. This process identifies a minimal set of high-quality candidates for final fine-grained matching, ensuring both high precision and computational efficiency.}
  \label{fig:visual}
\end{figure}


Despite these evident bottlenecks, the field has yet to fundamentally challenge the exhaustive scanning paradigm. 
Mainstream solutions~\cite{luo2022clip4clip, wu2023cap4video, yang2024dgl}, predominantly driven by Vision-Language Models (VLMs)~\cite{radford2021learning, li2023blip}, focus on optimizing joint embedding spaces to improve matching metrics. 
However, they inherently adhere to the $\mathcal{O}(N)$ brute-force search complexity. 
Moreover, achieving competitive performance with these methods typically relies on extensive task-specific fine-tuning, which limits their zero-shot generalization and applicability in open-world scenarios~\cite{wang2023videomae, wasim2023vita, ren2024timechat}.
Recently, emerging research has begun to explore Multimodal Large Language Models (MLLMs)~\cite{ye2026sugar, ye2025cat+} for deeper reasoning. 
While these generative approaches offer a compelling training-free solution with superior semantic understanding, applying them to large-scale retrieval exacerbates the efficiency-accuracy dilemma: running heavy reasoners on a full corpus is computationally prohibitive~\cite{li2024llama, yu2023self}. Furthermore, methods that rely on a single agent to parse content often suffer from \textit{Tunnel Vision}, where a solitary perspective overlooks subtle visual cues or hallucinates events due to cognitive overload~\cite{du2023improving}.
This poses a fundamental challenge: \textit{Can we synthesize the zero-shot reasoning capability of generative models with the scalability of retrieval systems, effectively bypassing the computational bottleneck of full-corpus traversal?}

To this end, we introduce \textbf{MAVIS}, a novel \textbf{M}ulti-\textbf{A}gent framework for \textbf{VI}deo \textbf{S}earch that redefines retrieval as a cooperative reasoning process. 
Instead of a mechanical brute-force scan, MAVIS operates like a team of human specialists tackling a complex investigation.
To enable efficient lookup, we first employ a \textit{Description Agent} to offload visual perception into a structured \textbf{Semantic Library}, effectively transforming raw pixels into a queryable textual index.
Upon receiving a user query, a \textit{Planner} first decomposes the complex intent into atomic semantic components (e.g., \textit{Action}, \textit{Object}, \textit{Scene}). 
Guided by this breakdown, MAVIS dynamically assembles a team of specialized agents, activating a dedicated ``expert'' only for the semantic perspectives explicitly present in the query.
In the retrieval phase, these agents work independently to nominate potential candidates based on their assigned focus.
Crucially, to ensure precision, they convene for a \textbf{Logic-aware Debate}: employing a strict veto protocol, the agents act as peer reviewers for one another, collectively rejecting any candidate that logically contradicts a specific expert's view.
This allows MAVIS to identify a compact set of \textit{``controversial''} candidates for final fine-grained verification, effectively realizing the philosophy of ``Inspect less, but understand more.''

In summary, our main contributions are:
\begin{itemize}
    \item We propose a \textbf{Structured Video Understanding} paradigm that explicitly parses videos into a semantic library, effectively bridges the granularity gap between dense video content and sparse textual queries.
    \item We design a \textbf{Collaborative Multi-Agent Framework} that dynamically assembles specialist agents based on query intent. We introduce a novel \textbf{Logic-aware Debate} where agents leverage a veto protocol to rigorously filter logical mismatches, significantly pruning the search space while preserving hard positives.
    \item Extensive experiments on MSR-VTT~\cite{xu2016msr}, MSVD~\cite{guadarrama2013youtube2text}, and ActivityNet~\cite{krishna2017dense} demonstrate that MAVIS turns generic pre-trained models into strong video retrievers, outperforming state-of-the-art fine-tuning methods.
\end{itemize}

\begin{figure*}[t]
  \centering
  \includegraphics[width=1.0\textwidth]{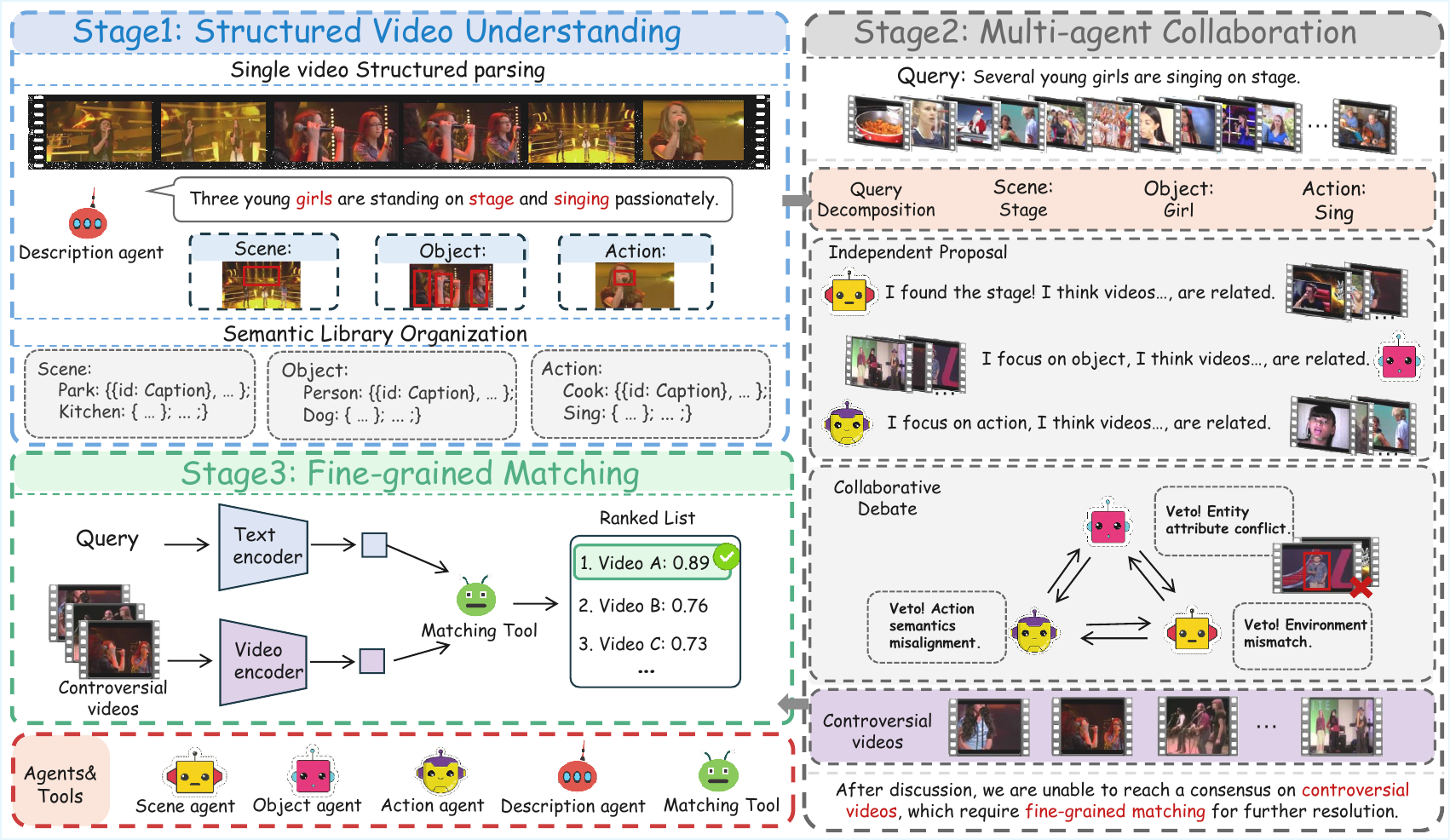}
  \caption{Overview of MAVIS. The pipeline consists of three progressive stages: (1) Structured Video Understanding: The Description Agent parses raw videos into structured semantic components (Scene, Object, Action) to construct a semantic library. (2) Multi-agent Collaboration: Given a user query, a planner decomposes the query. Role-specific agents independently propose candidates and engage in a collaborative debate to filter out mismatches. (3) Fine-grained Matching: For "controversial videos" where consensus is not reached, the Matching Tool utilizes fine-grained visual-text alignment to generate the final ranked list.}
  \label{fig:framework}
\end{figure*}

\section{Related Work}
\subsection{Text-to-video Retrieval}
Video retrieval faces significant challenges due to the modality gap between visual content and natural language~\cite{dong2022partially, zhang2025mgtr, zhang2026retrievingrecoverincompleteaudiovisual}. Traditional methods typically prioritize either retrieval accuracy through fine-grained cross-modal interaction~\cite{jin2023text, wu2025varcmp}, or inference efficiency by utilizing dual-encoder architectures~\cite{liu2025learning, fang2024linguistic}.
The emergence of large-scale Vision-Language Models (VLMs) such as CLIP~\cite{radford2021learning} has revolutionized video retrieval by enabling effective cross-modal alignment~\cite{tian2024holistic, liu2025learning, yang2024dgl}. However, deriving optimal performance typically requires extensive fine-tuning, inevitably escalating the computational overhead.
Recent research has also explored training-free approaches that leverage pre-trained models without fine-tuning. Systems like Merlin~\cite{han2024merlin} rely on multi-round interactions and iterative reasoning, enabling effective video-query matching without the need for model retraining. 
However, it still requires traversing the entire video library, relying on a single model through multiple rounds of interaction.
In contrast to these approaches, MAVIS avoids full-library traversal and leveraging multi-agent collaboration, leading to more efficient and accurate retrieval.

\subsection{Agent and tool use}
Recent advancements in large language models (LLMs) and multimodal LLMs (MLLMs) have significantly enhanced their reasoning and planning capabilities, driving the development of autonomous agents~\cite{zhang2024codeagent, luo2024video, ye2026eyes, ye2024cat}. 
Agent-based methods are effective at decomposing complex tasks~\cite{yao2022react, shen2023hugginggpt}, and the multi-agent paradigm has achieved significant success in various tasks, such as long video understanding~\cite{chen2025lvagent, yang2025vca}, where agent collaboration has substantially improved task comprehension and performance. 
Furthermore, by integrating external tools, these models have bridged the gap between general-purpose reasoning and real-world execution~\cite{fan2024videoagent, zhu2026H-GAR, zhu2026delta}. 
However, existing frameworks typically rely on heavy iterative reasoning or analyze videos individually, making them inefficient for retrieving targets from massive libraries. To Bridge this gap, we introduce MAVIS, a framework that repurposes multi-agent collaboration for efficient retrieval. MAVIS enhances retrieval accuracy specifically by leveraging agent consensus, while avoiding the computational burden of full-library traversal and model fine-tuning.

\section{Method}
\label{sec:method}
\subsection{Overview}
Given a natural language query $q$ and a video corpus $\mathcal{V} = \{v_i\}_{i=1}^{N}$, the goal of MAVIS is to retrieve the top-$k$ most relevant videos. Unlike traditional dual-encoder approaches, MAVIS adopts a \textit{coarse-to-fine} agentic workflow. The framework consists of three phases: (1) \textbf{Structured Video Understanding}, which transforms videos into semantic attributes; (2) \textbf{Collaborative Multi-Agent Pruning}, where specialized agents actively filter the corpus to identify a small set of "controversial" candidates; and (3) \textbf{Fine-grained Matching}, which performs fine-grained verification only on the survivors. The overall architecture is illustrated in Figure~\ref{fig:framework}.

\subsection{Structured Video Understanding}
Conventional video-text retrieval paradigms typically align holistic video and text embeddings within a joint latent space. However, this global alignment is inherently susceptible to \textbf{semantic asymmetry}.
As illustrated in Fig.~\ref{fig:visual}, a raw video naturally encapsulates exhaustive details, including concurrent entities, background events, and atmospheric nuances. 
In contrast, user queries are typically concise and intent-focused . 
This \textbf{granularity mismatch} allows extraneous visual details to act as noise in the embedding space, diluting the matching score for the core intent.
To address this, MAVIS introduces a structured video understanding process. Instead of ingesting noisy raw features, we leverage the visual reasoning capability of multimodal large language models (MLLMs) to actively \textit{filter} video content into explicit semantic attributes and concise descriptions. This process comprises two phases: \textit{Single Video Structured Parsing} and \textit{Semantic Library Organization}.

\vspace{0.1cm}
\noindent{\textbf{Single Video Structured Parsing.}}  
For each video $v_i$, we employ a pre-trained MLLM with a specialized prompt designed to suppress irrelevant details and extract only salient events. 
Specifically, the model is instructed to directly output a concise caption $c_i$ that mirrors the brevity of search queries, along with a structured semantic tuple:
\begin{equation}
    \mathbf{s}_i = \big( s_i^{\textsc{scn}}, s_i^{\textsc{obj}}, s_i^{\textsc{act}} \big)
\end{equation}
where each component contains standardized keywords (e.g., $s_i^{\textsc{obj}} = \{\text{cartoon character}\}$). 
By bypassing the generation of verbose descriptions, MAVIS ensures that the stored representation $c_i$ is semantically aligned with the granularity of potential text queries, minimizing the risk of length-induced mismatch.

\vspace{0.1cm}
\noindent{\textbf{Semantic Library Organization.}}
We reorganize the parsed corpus into a structured semantic library $\mathcal{L}$.
Given the open-vocabulary nature of MLLM outputs, raw tags often exhibit linguistic variations (e.g., \textit{``chatting''}, \textit{``communicating''}). To ensure retrieval robustness, we leverage the MLLM to normalize raw visual observations into a unified set of \textbf{Canonical Concepts} $\mathcal{K}$.
During the parsing phase, the model is instructed to consolidate diverse descriptive terms into standardized keywords.
This creates a coherent vocabulary $\mathcal{K}$ dynamically, ensuring that semantically identical content is indexed under the same key.

We structure the library into three domain-specific sub-libraries: $\mathcal{L}_{\textsc{scn}}$ (Scene), $\mathcal{L}_{\textsc{obj}}$ (Object), and $\mathcal{L}_{\textsc{act}}$ (Action).
Within each sub-library, videos are organized under specific semantic categories.
Formally, the sub-library $\mathcal{L}_{d}$ functions as a map where each key $k$ represents a specific concept, pointing to all relevant video instances:
\begin{equation}
    \mathcal{L}_{d}[k] = \big\{ (\text{id}=i, \text{cap}=c_i) \mid k \in s_i^{d} \big\},
\end{equation}
where $c_i$ denotes the concise caption.
This categorical organization empowers specialized agents to rapidly isolate relevant candidates by querying specific semantic keys, effectively converting the retrieval task from an exhaustive scan to a targeted lookup.

\subsection{Collaborative Multi-Agent Pruning}
\label{sec:collaboration}
\begin{figure}[t]                    
  \centering
  \includegraphics[width=1.0\columnwidth]{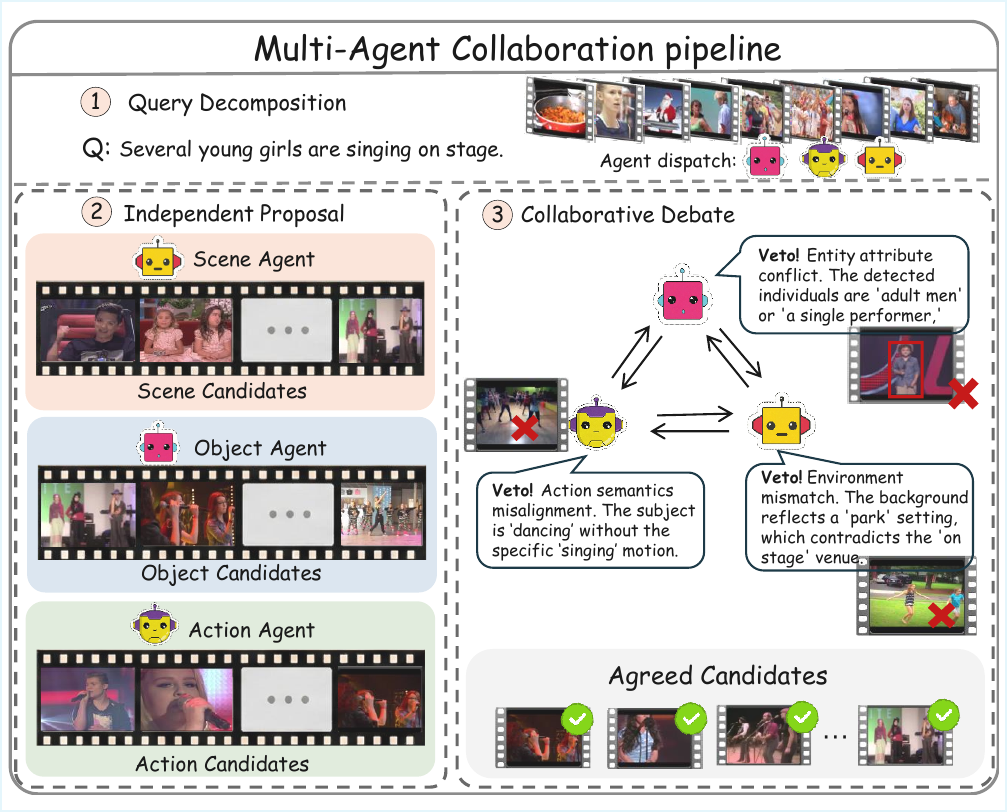}
  \caption{Multi-agent Collaboration pipeline: (1) Query Decomposition: Parsing the query into sub-tasks for agent dispatch. (2) Independent Proposal: Agents independently nominate initial candidates. (3) Collaborative Debate: A veto mechanism filters semantically inconsistent samples based on scene, object, and action constraints to reach a final consensus.}
  \label{fig:fig3}
\end{figure}

To avoid the prohibitive cost of full-corpus verification, MAVIS employs a multi-agent pruning stage. 
Drawing inspiration from human collaboration, MAVIS mirrors the workflow of expert teams: rather than exhaustively scrutinizing every document, it prioritizes efficiency by rapidly filtering candidates via domain-specific heuristics and reserving debate for high potential cases.
As illustrated in Fig.~\ref{fig:fig3}, we formalize this intuition into three step: \textit{Query Decomposition}, \textit{Independent Proposal}, and \textit{Collaborative Debate}.

\vspace{0.1cm}
\noindent{\textbf{Query Decomposition.}}
Upon receiving a query $q$, MAVIS first identifies the necessary semantic roles. A planner decomposes $q$ into a set of sub-intents $\mathcal{Q} = \{q^r \mid r \in \mathcal{R}\}$, where $\mathcal{R} \subseteq \{\textsc{scn}, \textsc{obj}, \textsc{act}\}$ represents the active semantic dimensions. For example, the query ``\textit{A dog running on the grass}'' activates all three agents, whereas ``\textit{A dog running}'' triggers the Object and Action agents ($A_{\textsc{obj}}$ and $A_{\textsc{act}}$). This selective activation ensures computational resources are not wasted on irrelevant dimensions.

\vspace{0.1cm}
\noindent{\textbf{Independent Proposal.}}
Each active agent $A_r$ acts as a domain specialist tasked with retrieving candidates from its dedicated sub-library $\mathcal{L}_r$.
Given the open-vocabulary discrepancy between the user's query $q$ and the stored canonical keys, the agent autonomously navigates the library index to identify the canonical category $k$ that semantically aligns with the intent, resolving linguistic variations.
Upon locating the correct category, the agent proceeds to verify the specific candidates. It acts as a strict evaluator, computing a confidence score $\phi_r(v_i)$ for each video by measuring the alignment between the query and the video's caption $C_{i,r}$:
\begin{equation}
    \phi_r(v_i) = \cos\left( E_{\text{VLM}}^{\text{txt}}(q^r), E_{\text{VLM}}^{\text{txt}}(C_{i,r}) \right),
\end{equation}
where $E_{\text{VLM}}^{\text{txt}}(\cdot)$ denotes the text encoder of a VLM, and $\cos(\cdot)$ computes the cosine similarity. This metric quantifies the agent's judgment confidence.
Based on this score, the agent generates a \textit{Proposal Message} $M_r$:
\begin{equation}
    M_r = \big\{ (v_i, \phi_r(v_i)) \mid \phi_r(v_i) > \tau_{\text{soft}} \big\},
\end{equation}
where $\tau_{\text{soft}}$ is a lenient threshold designed to maximize recall before entering the debate stage.

\vspace{0.1cm}
\noindent{\textbf{Collaborative Debate.}}
Instead of forcing agents to communicate point-to-point, the system serves as a central registry that asynchronously aggregates the independent proposal messages $M_r$ from all agents into a unified preliminary candidate pool $\mathcal{V}_{\text{pool}}$.
This union operation prioritizes recall, ensuring that a video proposed by \textit{any} domain expert is considered a potential candidate:
\begin{equation}
    \mathcal{V}_{\text{pool}} = \bigcup_{r \in \mathcal{R}} \{v_i \mid (v_i, \phi_r) \in M_r\}.
\end{equation}

To support a rigorous debate, the blackboard ensures that every agent has a ``vote'' on every candidate in $\mathcal{V}_{\text{pool}}$.
Since agents search independently, a video $v_i$ proposed by the Object agent might initially lack a score from the Action agent.
Therefore, for any $v_i \in \mathcal{V}_{\text{pool}}$, if an agent $A_r$ has not yet computed a score (i.e., $v_i \notin M_r$), it performs an on-demand check.
This synchronizes the blackboard, assigning a complete confidence vector $\Phi(v_i) = \{\phi_r(v_i) \mid r \in \mathcal{R}\}$ to every candidate.

With complete information available, the system applies a \textit{Veto Protocol} to strictly filter out mismatched videos.
The intuition is based on logical negation: a video remains valid only if \textit{no} domain expert identifies a hard conflict.
If the Object agent detects that a required entity is definitely missing ($\phi_{\textsc{obj}} < \tau_{\text{hard}}$), the video is discarded immediately, regardless of how well it matches the scene or action attributes.
Formally, the surviving set is derived by subtracting the vetoed candidates:
\begin{equation}
    \mathcal{V}_{\text{contro}} = \mathcal{V}_{\text{pool}} \setminus \big\{ v_i \in \mathcal{V}_{\text{pool}} \mid \exists r, \phi_r(v_i) < \tau_{\text{hard}} \big\},
\end{equation}
where $\tau_{\text{hard}}$ is a strict lower bound representing logical negation.
The remaining videos in $\mathcal{V}_{\text{contro}}$, termed \textit{Controversial Candidates}, are semantically plausible but require fine-grained verification in the next stage.

\begin{table*}[t]
\centering
\caption{Performance comparison on MSR-VTT, MSVD, and ActivityNet. We categorize methods into supervised fine-tuning, large-scale foundation models, and agentic methods. \textbf{MAVIS} achieves superior performance without training.}
\label{tab:performance_comparison}
\renewcommand{\arraystretch}{1.2}
\resizebox{1.0\textwidth}{!}{
\begin{tabular}{l|ccc|ccc|ccc}
\toprule
\multirow{2}{*}{\textbf{Method}} & \multicolumn{3}{c|}{\textbf{MSR-VTT}} & \multicolumn{3}{c|}{\textbf{MSVD}} & \multicolumn{3}{c}{\textbf{ActivityNet}} \\
& \textbf{R@1} & \textbf{R@5} & \textbf{R@10} & \textbf{R@1} & \textbf{R@5} & \textbf{R@10} & \textbf{R@1} & \textbf{R@5} & \textbf{R@10} \\
\midrule

\multicolumn{10}{l}{\cellcolor{gray!5} \textit{\textbf{Supervised / Fine-tuned Methods}}} \\
\quad CLIP4Clip~\cite{luo2022clip4clip} & 44.5 & 71.4 & 81.6 & 46.2 & 76.1 & 84.6 & 40.5 & 72.4 & 80.1 \\
\quad X-CLIP~\cite{ma2022x} & 49.3 & 75.8 & 84.8 & 50.4 & 80.6 & 87.1 & 46.4 & 75.9 & 84.5 \\
\quad Cap4Video~\cite{wu2023cap4video} & 51.4 & 75.7 & 83.9 & 51.8 & 80.8 & 88.3 & 49.2 & 77.3 & 85.1 \\
\quad TeachCLIP~\cite{tian2024holistic} & 46.8 & 74.3 & 82.6 & 47.4 & 77.3 & 84.5 & 42.2 & 72.7 & 84.0 \\
\quad T-MASS~\cite{wang2024text} & 52.7 & 77.1 & 85.6 & 53.0 & 81.0 & 88.5 & 48.5 & 76.4 & 84.2 \\
\midrule
\multicolumn{10}{l}{\cellcolor{gray!5} \textit{\textbf{Large-scale Video Foundation Models}}} \\
\quad VAST~\cite{chen2023vast} & 49.3 & 68.3 & 73.9 & 55.4 & 82.0 & 88.5 & 55.2 & 82.5 & 90.6 \\
\quad InternVideo2-6B~\cite{2025InternVideo2} & 55.9 & 78.3 & 85.1 & 59.3 & 84.4 & 89.6 & 63.2 & 85.6 & 92.5 \\
\quad LanguageBind-H~\cite{zhu2023languagebind} & 44.8 & 70.0 & 78.7 & 53.9 & 80.4 & 87.8 & 41.0 & 68.4 & 80.8 \\
\quad VideoPrism-g~\cite{2024VideoPrism} & 39.7 & 63.7 & - & 58.5 & 83.4 & 89.2 & 52.7 & 79.4 & - \\
\quad Marengo-2.6~\cite{twelvelabs2024marengo} & 49.35 & 73.47 & - & 61.0 & 85.3 & 91.5 & 55.36 & 82.55 & - \\
\midrule
\multicolumn{10}{l}{\cellcolor{gray!5} \textit{\textbf{Agentic Methods}}} \\
\quad MERLIN (Round 0)~\cite{han2024merlin}& 44.4 & 67.6 & 76.2 & 52.39 & 77.16 & 84.78 & 56.58 & 84.77 & 91.73 \\
\quad MERLIN (Round 3) & 72.6 & 91.8 & 95.6 & 71.79 & 91.79 & 96.87 & 66.05 & 90.97 & 95.54 \\
\quad MERLIN (Round 5) & 78.0 & 94.2 & \textbf{96.8} & 77.61 &\textbf{94.48} & 97.31 & 68.44 & 91.95 & 96.63 \\

\midrule
\rowcolor{gray!15} \quad \textbf{MAVIS (Ours)} & \textbf{78.6} & \textbf{94.9} & 96.3 & \textbf{78.1} & 94.25 & \textbf{97.33} & \textbf{69.15} & \textbf{92.4} & \textbf{96.8} \\
\bottomrule
\end{tabular}
}
\end{table*}

\subsection{Fine-grained Matching}
The previous collaborative pruning stage effectively filters out the vast majority of irrelevant videos. For the remaining candidates $\mathcal{V}_{\text{contro}}$, we employ a powerful Vision-Language Model to perform fine-grained verification. The final relevance score is computed as:
\begin{equation}
    S(q, v_i) = \text{Cos}\big( E_{\text{VLM}}^{\text{txt}}(q), E_{\text{VLM}}^{\text{vid}}(v_i) \big),
\end{equation}
where $E_{\text{VLM}}^{\text{txt}}(\cdot)$ and $E_{\text{VLM}}^{\text{vid}}(\cdot)$ denote the text and video encoders of the VLM, respectively. We rank $v_i \in \mathcal{V}_{\text{contro}}$ by $S(q, v_i)$ to return the top-$k$ results.
This strategy optimizes the efficiency-accuracy trade-off: by reserving VLM computations exclusively for the most plausible candidates, MAVIS achieves deep visual understanding at a fraction of the computational cost required for full-corpus scanning.

\section{Experiments}
\subsection{Datasets and Metrics}
We conduct our evaluations on three widely used video retrieval datasets: MSR-VTT~\cite{xu2016msr}, MSVD~\cite{guadarrama2013youtube2text}, and ActivityNet~\cite{krishna2017dense}. To ensure a strictly fair comparison, we adhere to the specific evaluation settings and data partitions provided by MERLIN~\cite{han2024merlin}. We evaluate performance using the standard recall at K (\textit{R@K}) metric, which measures the percentage of queries for which the correct video appears in the top \textit{K} retrieved results.


\subsection{Implementation Details}

For the initial structured parsing of the video corpus, we employ \textbf{Qwen3-omni-flash}\footnote{\url{https://github.com/QwenLM/Qwen3}} as the backbone for the Description Agent. This selection balances format accuracy with semantic alignment efficiency. The multi-agent retrieval team (Scene, Object, and Action) shares a unified reasoning backbone: \textbf{GPT-4o}\footnote{\url{https://openai.com/index/hello-gpt-4o/}}. We utilize role-specific system prompting to specialize this backbone into distinct domain experts.

For similarity scoring ($\phi_r$) during the proposal phase, we utilize OpenAI's \textbf{text-embedding-3-large} model\footnote{\url{https://platform.openai.com/docs/guides/embeddings}} to encode textual descriptions into high-dimensional vectors. For the fine-grained matching stage ($S(q, v_i)$), we employ \textbf{ViCLIP}~\cite{2023internvid} to encode the query and video frames into a shared embedding space for precise visual verification.

Regarding the threshold settings, we empirically set the lenient threshold $\tau_{\text{soft}} = 0.5$ to maximize recall in the proposal phase, ensuring potential candidates are not prematurely discarded. For the final selection and debate phase, we implement a strict confidence threshold of $\tau_{\text{hard}} = 0.3$ to enforce logical consistency through our veto protocol. We implement our framework using PyTorch. All experiments are conducted on a single NVIDIA A100 GPU to ensure computational consistency.

\subsection{Main Results}
We evaluate MAVIS across three video-text retrieval benchmarks: MSR-VTT, MSVD, and ActivityNet, as summarized in Table~\ref{tab:performance_comparison}. 
Compared to traditional supervised models, MAVIS demonstrates a substantial performance leap despite being entirely training-free. For instance, on MSR-VTT, MAVIS achieves 78.6\% R@1, which is considerably higher than fine-tuned models such as T-MASS (52.7\%) and Cap4Video (51.4\%).
While models like InternVideo2-6B and VideoPrism-g excel in general video understanding, their broad-scale training objectives may not be fully optimized for the fine-grained cross-modal alignment required in retrieval tasks. These results suggest that for retrieval-specific challenges, structured alignment strategies are more critical than purely scaling up model size, as MAVIS better filters irrelevant visual noise to match textual queries. In comparison with agentic baselines, MAVIS shows superior efficiency and accuracy. While MERLIN starts with relatively poor performance in Round 0 (e.g., 56.58\% R@1 on ActivityNet) and requires five iterative rounds to reach 68.44\%, MAVIS achieves 69.15\% R@1 in a single pass. This underscores the efficacy of our pipeline in handling temporal dependencies and filtering noise more efficiently than iterative refinement, while avoiding the heavy computational overhead of multiple reasoning rounds.

\subsection{Ablation Study}
To validate the contribution of each component in MAVIS, we conducted extensive ablation studies. We analyze the framework from three perspectives: (1) the effectiveness of the collaborative pruning mechanism, (2) the architectural design of agents and data representations, and (3) the trade-off between retrieval efficiency and accuracy.

\vspace{0.1cm}
\noindent{\textbf{Impact of Collaborative Debate.}} 
We first investigate the Logic-aware Debate mechanism (Table~\ref{tab:ablation_components}, Rows a-c). \textit{Union Only} (a) achieves high recall potential but suffers from sub-optimal precision and a prohibitively large candidate pool, as it fails to filter irrelevant videos proposed by independent agents. Conversely, \textit{Strict Intersection} (b) acts as an overly aggressive filter, significantly dropping R@1 to 58.4\% by missing valid candidates that lack full consensus among all agents. While \textit{Average Fusion} (c) serves as a strong baseline, it struggles with ``hard negatives'' where at least one specialized agent possesses a decisive veto signal. Our proposed \textbf{Union + Veto} strategy (MAVIS) achieves the optimal balance, outperforming Union Only by 3.4\% in R@1 while effectively pruning the candidate pool size from 68 to 23, ensuring high efficiency for the subsequent fine-grained stage.

\vspace{0.1cm}
\noindent{\textbf{Necessity of Agent Specialization.}}
Replacing our specialized agents with a \textit{Single General Agent} (Row d) leads to a notable 6.5\% drop in R@1. This performance decay suggests that a single agent suffers from increased cognitive load when parsing Composite queries, often leading to a ``semantic blurring'' effect where critical details are overshadowed by verbose context. In contrast, our multi agent design facilitates semantic isolation, enabling the Scene, Object, and Action agents to function as domain-specific experts. By isolating these semantic roles, each agent can selectively attend to the most discriminative cues.

\vspace{0.1cm}
\noindent{\textbf{Impact of Semantic Library.}}
Utilizing \textit{Raw Captions} directly (Row e) degrades performance significantly. This validates the necessity of our \textit{Structured Video Understanding} stage. Unlike free-form text which often contains redundant noise or loosely coupled descriptions, structured parsing provides explicit semantic guidance by routing categorical attributes to their corresponding agents. This modular specialization allows each agent to focus exclusively on verified attributes within its domain, effectively preventing "semantic bleeding" and minimizing hallucinations.
\begin{table}[!t]
\centering
\small
\caption{Comprehensive ablation study on MAVIS components in MSR-VTT. $|\mathcal{V}_{\text{contro}}|$ denotes the average number of videos retained per query for the fine-grained stage. The default MAVIS setting is highlighted in \textbf{gray}.}
\label{tab:ablation_components}
\renewcommand{\arraystretch}{1.3} 
\setlength{\tabcolsep}{6pt}
\resizebox{1.0\columnwidth}{!}{
\begin{tabular}{l|ccc|c}
\toprule
\textbf{Method Setting} & \textbf{R@1} & \textbf{R@5} & \textbf{R@10} & \textbf{$|\mathcal{V}_{\text{contro}}|$} \\
\midrule
\rowcolor{gray!15} \textbf{MAVIS (Full Model)} & \textbf{78.6} & \textbf{94.9} & \textbf{96.3} & \textbf{16} \\
\midrule
\multicolumn{5}{l}{\textit{Q1: Impact of Collaborative Debate Strategy}} \\
\quad (a) Union Only (No Veto) & 75.2 & 90.1 & 92.5 & 58 \\
\quad (b) Strict Intersection ($\cap$) & 58.4 & 72.3 & 78.2 & \textbf{8} \\
\quad (c) Average Fusion Score & 76.8 & 92.5 & 94.1 & 31 \\
\midrule
\multicolumn{5}{l}{\textit{Q2: Necessity of Agent Specialization}} \\
\quad (d) Single General Agent & 72.1 & 88.4 & 90.8 & 57 \\
\midrule
\multicolumn{5}{l}{\textit{Q3: Impact of Semantic Library}} \\
\quad (e) w/o Structured Attributes & 70.5 & 84.2 & 87.5 & 69 \\
\bottomrule
\end{tabular}
}
\end{table}

\begin{table}[!t]
\centering
\small
\caption{Efficiency-Accuracy Trade-off analysis on MSR-VTT. FLOPs are estimated for the inference of a single query against the full corpus. Speedup is calculated relative to the Fine Only (ViCLIP).}
\label{tab:efficiency_tradeoff}
\renewcommand{\arraystretch}{1.53}
\resizebox{1.0\columnwidth}{!}{
\begin{tabular}{l|ccc|cc}
\toprule
\textbf{Paradigm} & \textbf{R@1} & \textbf{R@5} & \textbf{R@10} & \textbf{FLOPs (G)} & \textbf{Speedup} \\
\midrule
(i) Coarse Only (Agent debate) & 68.0 & 82.5 & 88.4 & \textbf{1} & \textbf{100,000$\times$} \\
(ii) Fine Only (Brute-force ViCLIP) & 50.8 & 74.8 & 82.9 & 100,000 & 1$\times$ \\
\midrule
\rowcolor{gray!15} \textbf{(iii) MAVIS (Coarse-to-Fine)} & 78.6 & 94.9 & 96.3 & 3200 & 31.2$\times$ \\
\bottomrule
\end{tabular}
}
\end{table}

\vspace{0.1cm} 
\noindent{\textbf{Efficiency-Accuracy Trade-off.}}
We evaluate the ``Inspect less, understand more'' philosophy in Table~\ref{tab:efficiency_tradeoff}. (i) \textbf{Coarse Only} retrieval (Agent debate) is extremely efficient, yielding a 100,000$\times$ speedup by matching structured captions in the text domain; however, it lacks final visual confirmation. (ii) \textbf{Fine Only (Brute-force ViCLIP)} represents the standard zero-shot visual baseline. Due to the lack of structured semantic reasoning, it achieves a lower precision while incurring a prohibitive computational cost of 100,000G FLOPs. (iii) \textbf{MAVIS} successfully bridges this gap and significantly outperforms the brute-force baseline. By leveraging the collaborative debate to prune the search space to just 16 candidates, MAVIS achieves a peak performance of \textbf{78.6\% R@1}. This demonstrates that our collaborative pruning and reasoning mechanism is not only efficient but also essential for resolving complex semantic queries that monolithic VLMs struggle to handle.


\begin{table}[t]
\centering
\small
\caption{Ablation study of thresholds $\tau_{\text{soft}}$ and $\tau_{\text{hard}}$ on the MSR-VTT dataset. $|\mathcal{V}_{\text{contro}}|$ denotes the average number of candidates requiring fine-grained matching.}
\label{tab:ablation_thresholds}
\setlength{\tabcolsep}{8pt}  
\begin{tabular}{ccccc}
\toprule
$\tau_{\text{soft}}$ & $\tau_{\text{hard}}$ & $|\mathcal{V}_{\text{contro}}|$ & R@1 ($\uparrow$) & R@10 ($\uparrow$) \\ \midrule
0.3 & 0.3 & 42.1 & 78.8 & 96.2 \\
\rowcolor{gray!10} \textbf{0.5} & \textbf{0.3} & \textbf{16.4} & \textbf{78.6} & \textbf{96.3} \\
0.7 & 0.3 & 5.8 & 72.4 & 85.5 \\ \midrule
0.5 & 0.1 & 38.2 & 78.7 & 96.5 \\
\rowcolor{gray!10} \textbf{0.5} & \textbf{0.3} & \textbf{16.4} & \textbf{78.6} & \textbf{96.3} \\
0.5 & 0.5 & 4.2 & 65.1 & 82.4 \\ \bottomrule
\end{tabular}
\end{table}
\noindent{\textbf{Ablation on Retrieval Thresholds.}}
We evaluate the sensitivity of MAVIS to the proposal threshold $\tau_{\text{soft}}$ and the veto threshold $\tau_{\text{hard}}$ in Table~\ref{tab:ablation_thresholds}. 
The lenient threshold $\tau_{\text{soft}}$ is designed to maximize the coverage of the preliminary candidate pool $\mathcal{V}_{\text{pool}}$. While $\tau_{\text{soft}}=0.3$ slightly boosts recall, it bloats the initial candidate pool to 42.1, increasing computational overhead; $\tau_{\text{soft}}=0.5$ serves as the optimal filter to maintain a manageable pool size.

The strict threshold $\tau_{\text{hard}}$ represents the boundary for logical negation within our collaborative veto protocol.
When $\tau_{\text{hard}}$ is set too high, the system becomes overly aggressive, leading to ``over-vetoing'' where valid positives are discarded due to minor semantic discrepancies.
Conversely, a lower $\tau_{\text{hard}}$ weakens the logical filter, failing to eliminate irrelevant distractors and resulting in a larger $\mathcal{V}_{\text{contro}}$. 

\subsection{Case Study}
\begin{figure}[t] 
    \centering
    \includegraphics[width=\columnwidth]{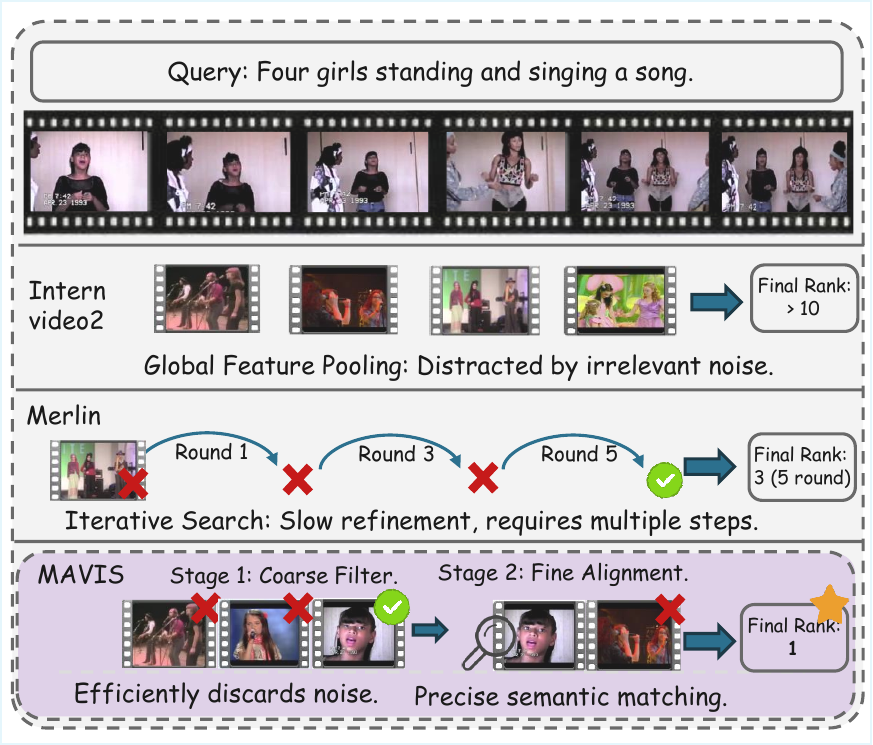}
    \caption{\textbf{Qualitative comparison of MAVIS, InternVideo2, and Merlin.} Case A: InternVideo2 suffers from retrieval biases as its global perception over-fits to background context rather than specific query intents. Case B: Merlin relies on sequential multi-round refinement. MAVIS successfully retrieves the ground truth via its logic-aware veto protocol.}
   
    \label{fig:case_study}
\end{figure}

To qualitatively evaluate the reasoning capabilities of MAVIS, we perform a comparative analysis against two representative paradigms: \textbf{InternVideo2}, a perception foundation model, and \textbf{Merlin}, an MLLM-based multi-round reasoning agent. As illustrated in Figure~\ref{fig:case_study}.
InternVideo2 relies on global embeddings to match video features with textual queries. In cases of \textit{semantic asymmetry}, InternVideo2 often yields false positives. The dominant scene features statistically overwhelm subtle object cues in the global latent space, leading to a loss of fine-grained precision.
Merlin utilizes an MLLM to orchestrate external tools through multiple rounds of reasoning. However, it is prone to \textit{tunnel vision} and reasoning hallucinations due to its single-agent architecture. 

MAVIS resolves these challenges by replacing holistic matching with a collaborative veto protocol. By assigning specialized agents to verify Scene, Object, and Action dimensions independently, MAVIS ensures that any semantic mismatch triggers an immediate veto, effectively eliminating hallucinations and perception drowning.

\section{Conclusion}
We introduced MAVIS, a training-free framework for text-to-video retrieval that replaces exhaustive full-corpus matching with a structured, coarse-to-fine retrieval paradigm. By first transforming raw videos into a semantic library and then performing retrieval through query decomposition, specialized agent collaboration, and logic-aware debate, MAVIS effectively addresses the granularity mismatch between sparse textual queries and information-dense video content. This design allows the system to prune large amounts of irrelevant information before invoking expensive fine-grained matching, thereby improving both retrieval efficiency and semantic precision.

Experiments on MSR-VTT, MSVD, and ActivityNet show that MAVIS achieves strong retrieval performance without task-specific fine-tuning, while maintaining a favorable efficiency--accuracy trade-off. The ablation studies further verify that the structured semantic library, agent specialization, and veto-based collaborative pruning are all important to the overall effectiveness of the framework. Taken together, these results suggest that cooperative multi-agent reasoning is a promising alternative to conventional embedding-based retrieval, and may provide a scalable and interpretable direction for future multimodal retrieval systems.

\section{Limitations}
Despite its promising results, MAVIS has several limitations. First, as a training-free framework, its performance is bounded by the zero-shot perception and reasoning capabilities of the underlying foundation models. Errors introduced during the structured parsing stage, such as omitted entities, inaccurate actions, or hallucinated descriptions, may propagate through the retrieval pipeline, since the current framework does not include an explicit mechanism for correcting upstream mistakes.

Second, MAVIS relies on empirically chosen proposal and veto thresholds, and its retrieval quality is therefore sensitive to threshold calibration. Although the selected settings perform well on the evaluated benchmarks, they may require adjustment in domains with different semantic distributions or higher visual ambiguity. In addition, the current agent design mainly focuses on scene, object, and action, which may be insufficient for more complex retrieval scenarios involving temporal dynamics, audio cues, OCR, or fine-grained relational reasoning. Finally, because MAVIS builds on off-the-shelf MLLMs and VLMs, it may also inherit their biases and failure modes, which should be considered when deploying the framework in real-world applications.

\section*{Ethics Statement}

In accordance with the ACL Ethics Policy, we provide the following statement regarding our work.
We utilized an AI-based language model exclusively for rephrasing and polishing the textual content of this paper to improve its clarity and linguistic quality. We explicitly state that the core methodology, experimental designs, and empirical results presented in this work are entirely original and were not generated or altered by any AI assistant. 
MAVIS is a training-free framework that leverages off-the-shelf Multimodal Large Language Models (MLLMs), such as GPT-4o and Qwen3. These large-scale models may reflect societal biases (e.g., gender, age, or race) present in their massive pre-training corpora. While MAVIS introduces a logic-aware veto protocol to enhance retrieval precision, it may still inherit or propagate such biases during the semantic parsing or scoring phases. Users should be cautious when deploying our framework in sensitive real-world applications.

\bibliography{custom}

@inproceedings{xu2016msr,
  title={Msr-vtt: A large video description dataset for bridging video and language},
  author={Xu, Jun and Mei, Tao and Yao, Ting and Rui, Yong},
  booktitle={Proceedings of the IEEE conference on computer vision and pattern recognition},
  pages={5288--5296},
  year={2016}
}

@inproceedings{guadarrama2013youtube2text,
  title={Youtube2text: Recognizing and describing arbitrary activities using semantic hierarchies and zero-shot recognition},
  author={Guadarrama, Sergio and Krishnamoorthy, Niveda and Malkarnenkar, Girish and Venugopalan, Subhashini and Mooney, Raymond and Darrell, Trevor and Saenko, Kate},
  booktitle={Proceedings of the IEEE international conference on computer vision},
  pages={2712--2719},
  year={2013}
}

@inproceedings{radford2021learning,
  title={Learning transferable visual models from natural language supervision},
  author={Radford, Alec and Kim, Jong Wook and Hallacy, Chris and Ramesh, Aditya and Goh, Gabriel and Agarwal, Sandhini and Sastry, Girish and Askell, Amanda and Mishkin, Pamela and Clark, Jack and others},
  booktitle={International conference on machine learning},
  pages={8748--8763},
  year={2021},
  organization={PmLR}
}

@article{luo2022clip4clip,
  title={Clip4clip: An empirical study of clip for end to end video clip retrieval and captioning},
  author={Luo, Huaishao and Ji, Lei and Zhong, Ming and Chen, Yang and Lei, Wen and Duan, Nan and Li, Tianrui},
  journal={Neurocomputing},
  volume={508},
  pages={293--304},
  year={2022},
  publisher={Elsevier}
}

@inproceedings{jin2023text,
  title={Text-video retrieval with disentangled conceptualization and set-to-set alignment},
  author={Jin, Peng and Li, Hao and Cheng, Zesen and Huang, Jinfa and Wang, Zhennan and Yuan, Li and Liu, Chang and Chen, Jie},
  booktitle={Proceedings of the Thirty-Second International Joint Conference on Artificial Intelligence},
  pages={938--946},
  year={2023}
}

@article{2023InternVid,
  title={InternVid: A Large-scale Video-Text Dataset for Multimodal Understanding and Generation},
  author={ Wang, Yi  and  He, Yinan  and  Li, Yizhuo  and  Li, Kunchang  and  Yu, Jiashuo  and  Ma, Xin  and  Li, Xinhao  and  Chen, Guo  and  Chen, Xinyuan  and  Wang, Yaohui },
  year={2023},
}

@inproceedings{tian2024holistic,
  title={Holistic features are almost sufficient for text-to-video retrieval},
  author={Tian, Kaibin and Zhao, Ruixiang and Xin, Zijie and Lan, Bangxiang and Li, Xirong},
  booktitle={Proceedings of the IEEE/CVF conference on computer vision and pattern recognition},
  pages={17138--17147},
  year={2024}
}

@article{han2024merlin,
  title={Merlin: Multimodal embedding refinement via llm-based iterative navigation for text-video retrieval-rerank pipeline},
  author={Han, Donghoon and Park, Eunhwan and Lee, Gisang and Lee, Adam and Kwak, Nojun},
  journal={arXiv preprint arXiv:2407.12508},
  year={2024}
}

@inproceedings{wang2024text,
  title={Text is mass: Modeling as stochastic embedding for text-video retrieval},
  author={Wang, Jiamian and Sun, Guohao and Wang, Pichao and Liu, Dongfang and Dianat, Sohail and Rabbani, Majid and Rao, Raghuveer and Tao, Zhiqiang},
  booktitle={Proceedings of the IEEE/CVF conference on computer vision and pattern recognition},
  pages={16551--16560},
  year={2024}
}

@inproceedings{yang2024dgl,
  title={Dgl: Dynamic global-local prompt tuning for text-video retrieval},
  author={Yang, Xiangpeng and Zhu, Linchao and Wang, Xiaohan and Yang, Yi},
  booktitle={Proceedings of the AAAI Conference on Artificial Intelligence},
  volume={38},
  number={7},
  pages={6540--6548},
  year={2024}
}

@inproceedings{wu2023cap4video,
  title={Cap4video: What can auxiliary captions do for text-video retrieval?},
  author={Wu, Wenhao and Luo, Haipeng and Fang, Bo and Wang, Jingdong and Ouyang, Wanli},
  booktitle={Proceedings of the IEEE/CVF conference on computer vision and pattern recognition},
  pages={10704--10713},
  year={2023}
}

@inproceedings{wang2023videomae,
  title={Videomae v2: Scaling video masked autoencoders with dual masking},
  author={Wang, Limin and Huang, Bingkun and Zhao, Zhiyu and Tong, Zhan and He, Yinan and Wang, Yi and Wang, Yali and Qiao, Yu},
  booktitle={Proceedings of the IEEE/CVF conference on computer vision and pattern recognition},
  pages={14549--14560},
  year={2023}
}

@inproceedings{dong2022partially,
  title={Partially relevant video retrieval},
  author={Dong, Jianfeng and Chen, Xianke and Zhang, Minsong and Yang, Xun and Chen, Shujie and Li, Xirong and Wang, Xun},
  booktitle={Proceedings of the 30th ACM International Conference on Multimedia},
  pages={246--257},
  year={2022}
}

@inproceedings{wasim2023vita,
  title={Vita-clip: Video and text adaptive clip via multimodal prompting},
  author={Wasim, Syed Talal and Naseer, Muzammal and Khan, Salman and Khan, Fahad Shahbaz and Shah, Mubarak},
  booktitle={Proceedings of the IEEE/CVF Conference on Computer Vision and Pattern Recognition},
  pages={23034--23044},
  year={2023}
}

@inproceedings{ren2024timechat,
  title={Timechat: A time-sensitive multimodal large language model for long video understanding},
  author={Ren, Shuhuai and Yao, Linli and Li, Shicheng and Sun, Xu and Hou, Lu},
  booktitle={Proceedings of the IEEE/CVF Conference on Computer Vision and Pattern Recognition},
  pages={14313--14323},
  year={2024}
}

@inproceedings{li2024llama,
  title={Llama-vid: An image is worth 2 tokens in large language models},
  author={Li, Yanwei and Wang, Chengyao and Jia, Jiaya},
  booktitle={European Conference on Computer Vision},
  pages={323--340},
  year={2024},
  organization={Springer}
}

@article{yu2023self,
  title={Self-chained image-language model for video localization and question answering},
  author={Yu, Shoubin and Cho, Jaemin and Yadav, Prateek and Bansal, Mohit},
  journal={Advances in Neural Information Processing Systems},
  volume={36},
  pages={76749--76771},
  year={2023}
}

@inproceedings{li2023blip,
  title={Blip-2: Bootstrapping language-image pre-training with frozen image encoders and large language models},
  author={Li, Junnan and Li, Dongxu and Savarese, Silvio and Hoi, Steven},
  booktitle={International conference on machine learning},
  pages={19730--19742},
  year={2023},
  organization={PMLR}
}

@inproceedings{du2023improving,
  title={Improving factuality and reasoning in language models through multiagent debate},
  author={Du, Yilun and Li, Shuang and Torralba, Antonio and Tenenbaum, Joshua B and Mordatch, Igor},
  booktitle={Forty-first International Conference on Machine Learning},
  year={2023}
}

@inproceedings{wu2025varcmp,
  title={VarCMP: Adapting Cross-Modal Pre-Training Models for Video Anomaly Retrieval},
  author={Wu, Peng and Su, Wanshun and He, Xiangteng and Wang, Peng and Zhang, Yanning},
  booktitle={Proceedings of the AAAI Conference on Artificial Intelligence},
  volume={39},
  number={8},
  pages={8423--8431},
  year={2025}
}

@inproceedings{liu2025learning,
  title={Learning Dynamic Similarity by Bidirectional Hierarchical Sliding Semantic Probe for Efficient Text Video Retrieval},
  author={Liu, Yang and Huang, Shudong and Xiong, Deng and Lv, Jiancheng},
  booktitle={Proceedings of the AAAI Conference on Artificial Intelligence},
  volume={39},
  number={6},
  pages={5667--5675},
  year={2025}
}

@inproceedings{yang2025vca,
  title={Vca: Video curious agent for long video understanding},
  author={Yang, Zeyuan and Chen, Delin and Yu, Xueyang and Shen, Maohao and Gan, Chuang},
  booktitle={Proceedings of the IEEE/CVF International Conference on Computer Vision},
  pages={20168--20179},
  year={2025}
}

@inproceedings{yao2022react,
  title={React: Synergizing reasoning and acting in language models},
  author={Yao, Shunyu and Zhao, Jeffrey and Yu, Dian and Du, Nan and Shafran, Izhak and Narasimhan, Karthik R and Cao, Yuan},
  booktitle={The eleventh international conference on learning representations}
}

@article{shen2023hugginggpt,
  title={Hugginggpt: Solving ai tasks with chatgpt and its friends in hugging face},
  author={Shen, Yongliang and Song, Kaitao and Tan, Xu and Li, Dongsheng and Lu, Weiming and Zhuang, Yueting},
  journal={Advances in Neural Information Processing Systems},
  volume={36},
  pages={38154--38180},
  year={2023}
}

@inproceedings{krishna2017dense,
  title={Dense-captioning events in videos},
  author={Krishna, Ranjay and Hata, Kenji and Ren, Frederic and Fei-Fei, Li and Carlos Niebles, Juan},
  booktitle={Proceedings of the IEEE international conference on computer vision},
  pages={706--715},
  year={2017}
}

@inproceedings{2025InternVideo2,
  title={InternVideo2: Scaling Foundation Models forMultimodal Video Understanding},
  author={ Wang, Yi  and  Li, Kunchang  and  Li, Xinhao  and  Yu, Jiashuo  and  He, Yinan  and  Chen, Guo  and  Pei, Baoqi  and  Zheng, Rongkun  and  Wang, Zun  and  Shi, Yansong },
  booktitle={European Conference on Computer Vision},
  year={2025},
}

@inproceedings{tang2025muse,
  title={Muse: Mamba is efficient multi-scale learner for text-video retrieval},
  author={Tang, Haoran and Cao, Meng and Huang, Jinfa and Liu, Ruyang and Jin, Peng and Li, Ge and Liang, Xiaodan},
  booktitle={Proceedings of the AAAI Conference on Artificial Intelligence},
  volume={39},
  number={7},
  pages={7238--7246},
  year={2025}
}

@inproceedings{wang2025koala,
  title={Koala-36m: A large-scale video dataset improving consistency between fine-grained conditions and video content},
  author={Wang, Qiuheng and Shi, Yukai and Ou, Jiarong and Chen, Rui and Lin, Ke and Wang, Jiahao and Jiang, Boyuan and Yang, Haotian and Zheng, Mingwu and Tao, Xin and others},
  booktitle={Proceedings of the Computer Vision and Pattern Recognition Conference},
  pages={8428--8437},
  year={2025}
}

@inproceedings{ma2022x,
  title={X-clip: End-to-end multi-grained contrastive learning for video-text retrieval},
  author={Ma, Yiwei and Xu, Guohai and Sun, Xiaoshuai and Yan, Ming and Zhang, Ji and Ji, Rongrong},
  booktitle={Proceedings of the 30th ACM international conference on multimedia},
  pages={638--647},
  year={2022}
}

@article{chen2023vast,
  title={Vast: A vision-audio-subtitle-text omni-modality foundation model and dataset},
  author={Chen, Sihan and Li, Handong and Wang, Qunbo and Zhao, Zijia and Sun, Mingzhen and Zhu, Xinxin and Liu, Jing},
  journal={Advances in Neural Information Processing Systems},
  volume={36},
  pages={72842--72866},
  year={2023}
}

@article{zhu2023languagebind,
  title={Languagebind: Extending video-language pretraining to n-modality by language-based semantic alignment},
  author={Zhu, Bin and Lin, Bin and Ning, Munan and Yan, Yang and Cui, Jiaxi and Wang, HongFa and Pang, Yatian and Jiang, Wenhao and Zhang, Junwu and Li, Zongwei and others},
  journal={arXiv preprint arXiv:2310.01852},
  year={2023}
}

@misc{twelvelabs2024marengo,
  author = {{Twelve Labs}},
  title = {Introducing {Marengo 2.6}: A State-of-the-Art Video Foundation Model for Any-to-Any Search},
  year = {2024},
  howpublished = {\url{https://www.twelvelabs.io/blog/introducing-marengo-2-6}},
  note = {Accessed: 2026-01-05}
}

@article{2024VideoPrism,
  title={VideoPrism: A Foundational Visual Encoder for Video Understanding},
  author={ Zhao, Long  and  Gundavarapu, Nitesh B.  and  Yuan, Liangzhe  and  Zhou, Hao  and  Yan, Shen  and  Sun, Jennifer J.  and  Friedman, Luke  and  Qian, Rui  and  Weyand, Tobias  and  Zhao, Yue },
  year={2024},
}

@article{fang2024linguistic,
  title={Linguistic hallucination for text-based video retrieval},
  author={Fang, Sheng and Dang, Tiantian and Wang, Shuhui and Huang, Qingming},
  journal={IEEE Transactions on Circuits and Systems for Video Technology},
  year={2024},
  publisher={IEEE}
}

@article{zhang2024codeagent,
  title={Codeagent: Enhancing code generation with tool-integrated agent systems for real-world repo-level coding challenges},
  author={Zhang, Kechi and Li, Jia and Li, Ge and Shi, Xianjie and Jin, Zhi},
  journal={arXiv preprint arXiv:2401.07339},
  year={2024}
}

@article{chen2025lvagent,
  title={Lvagent: Long video understanding by multi-round dynamical collaboration of mllm agents},
  author={Chen, Boyu and Yue, Zhengrong and Chen, Siran and Wang, Zikang and Liu, Yang and Li, Peng and Wang, Yali},
  journal={arXiv preprint arXiv:2503.10200},
  year={2025}
}

@article{luo2024video,
  title={Video-rag: Visually-aligned retrieval-augmented long video comprehension},
  author={Luo, Yongdong and Zheng, Xiawu and Yang, Xiao and Li, Guilin and Lin, Haojia and Huang, Jinfa and Ji, Jiayi and Chao, Fei and Luo, Jiebo and Ji, Rongrong},
  journal={arXiv preprint arXiv:2411.13093},
  year={2024}
}

@inproceedings{fan2024videoagent,
  title={Videoagent: A memory-augmented multimodal agent for video understanding},
  author={Fan, Yue and Ma, Xiaojian and Wu, Rujie and Du, Yuntao and Li, Jiaqi and Gao, Zhi and Li, Qing},
  booktitle={European Conference on Computer Vision},
  pages={75--92},
  year={2024},
  organization={Springer}
}

@inproceedings{zhu2026H-GAR,
  title={H-GAR: A Hierarchical Interaction Framework via Goal-Driven Observation-Action Refinement for Robotic Manipulation},
  author={Zhu, Yijie and Shao, Rui and Liu, Ziyang and He, Jie and Liu, Jizhihui and Wang, Jiuru and Yu, Zitong},
  booktitle={Proceedings of the AAAI Conference on Artificial Intelligence},
  year={2026}
}

@article{zhu2026delta,
  title={$\delta$ VLA: Prior-Guided Vision-Language-Action Models via World Knowledge Variation},
  author={Zhu, Yijie and He, Jie and Shao, Rui and Yuan, Kaishen and Tan, Tao and Yuan, Xiaochen and Yu, Zitong},
  journal={arXiv preprint arXiv:2603.08361},
  year={2026}
}

@inproceedings{ye2024cat,
  title={Cat: Enhancing multimodal large language model to answer questions in dynamic audio-visual scenarios},
  author={Ye, Qilang and Yu, Zitong and Shao, Rui and Xie, Xinyu and Torr, Philip and Cao, Xiaochun},
  booktitle={European Conference on Computer Vision},
  pages={146--164},
  year={2024},
  organization={Springer}
}

@article{ye2025cat+,
  title={Cat+: Investigating and enhancing audio-visual understanding in large language models},
  author={Ye, Qilang and Yu, Zitong and Shao, Rui and Cui, Yawen and Kang, Xiangui and Liu, Xin and Torr, Philip and Cao, Xiaochun},
  journal={IEEE Transactions on Pattern Analysis and Machine Intelligence},
  year={2025},
  publisher={IEEE}
}

@inproceedings{ye2026eyes,
  title={When Eyes and Ears Disagree: Can MLLMs Discern Audio-Visual Confusion?},
  author={Ye, Qilang and Zeng, Wei and Liu, Meng and Zhang, Jie and Hu, Yupeng and Yu, Zitong and Zhou, Yu},
  booktitle={Proceedings of the AAAI Conference on Artificial Intelligence},
  volume={40},
  number={14},
  pages={11955--11963},
  year={2026}
}

@inproceedings{ye2026sugar,
  title={SUGAR: Learning Skeleton Representation with Visual-Motion Knowledge for Action Recognition},
  author={Ye, Qilang and Zhou, Yu and He, Lian and Zhang, Jie and Guo, Xuanming and Zhang, Jiayu and Tan, Mingkui and Xie, Weicheng and Sun, Yue and Tan, Tao and others},
  booktitle={Proceedings of the AAAI Conference on Artificial Intelligence},
  volume={40},
  number={21},
  pages={17930--17938},
  year={2026}
}

@misc{zhang2026retrievingrecoverincompleteaudiovisual,
      title={Retrieving to Recover: Towards Incomplete Audio-Visual Question Answering via Semantic-consistent Purification}, 
      author={Jiayu Zhang and Shuo Ye and Qilang Ye and Zihan Song and Jiajian Huang and Zitong Yu},
      year={2026},
      eprint={2604.10695},
      archivePrefix={arXiv},
      primaryClass={cs.CV},
      url={https://arxiv.org/abs/2604.10695}, 
}

@article{zhang2025mgtr,
  title={Mgtr-miss: More ground truth retrieving based multimodal interaction and semantic supervision for video description},
  author={Zhang, Jiayu and Tang, Pengjie and Tan, Yunlan and Wang, Hanli},
  journal={Neural Networks},
  pages={107817},
  year={2025},
  publisher={Elsevier}
}

\appendix
\section{Appendix: Implementation Details}
\noindent{\textbf{model setup.}}
The performance of MAVIS heavily relies on the quality of the structured captions generated in the first stage. To select the optimal backbone for the Description Agent, we conducted a preliminary study comparing state-of-the-art Multimodal Large Language Models (MLLMs), including \textbf{GPT-4.1}, \textbf{Gemini-2.5-Pro}, and \textbf{Qwen3-omni-flash}. We constructed a dataset of 200 diverse videos sampled from the training set of MSR-VTT. 
For each model, we evaluated its instruction-following capability (i.e., adherence to the JSON output format) and the semantic accuracy of the extracted attributes. Specifically, we employ the \textit{Caption-Query Alignment Score} as our evaluation metric, which measures the retrieval recall of the generated captions against ground-truth queries
using OpenAI's \textbf{text-embedding-3-large} model. The motivation behind this metric is to ensure that the generated captions are highly semantic-aligned with potential user queries, thereby bridging the modality gap more effectively during the retrieval phase. 
As shown in Table~\ref{tab:model_selection}, \textbf{Qwen3-omni-flash} achieves the highest scores in both format accuracy and semantic alignment. Consequently, we adopted Qwen3-omni-flash as the backbone for the Description Agent to construct the Semantic Library.

\begin{figure*}[t] 
    \centering
    
    \includegraphics[width=1.0\textwidth]{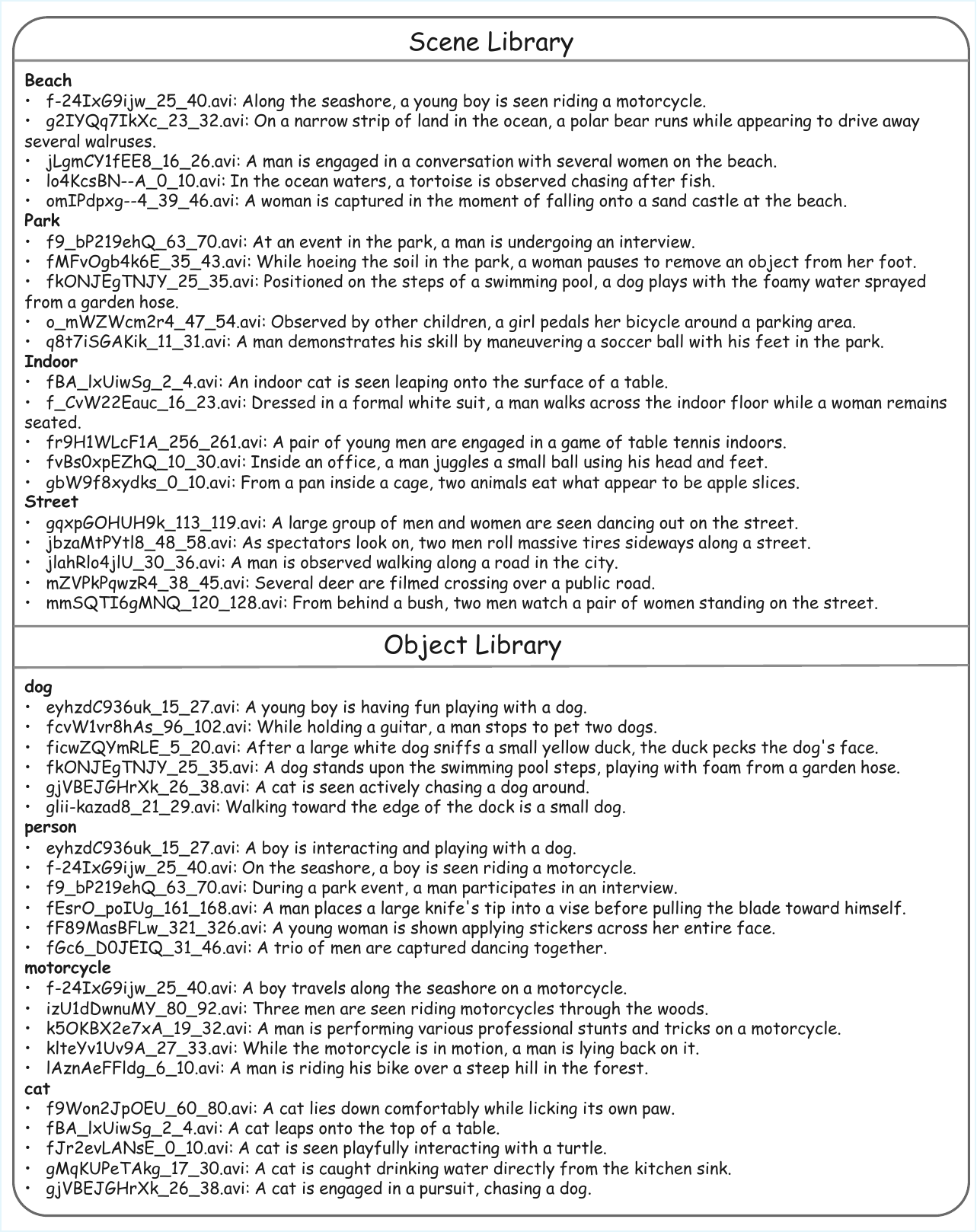} 
    \caption{\textbf{Visualization of Semantic Library.} This figure illustrates samples from sub-libraries. Each entry maps a specific video ID to its corresponding concise semantic summary. This structured organization enables the Planner and specialized agents to perform targeted lookups rather than exhaustive scans.}   
    \label{fig:library_content}
\end{figure*}


\section{visualization of Semantic Library}

The MAVIS semantic library is organized as a structured inverted index, designed to replace conventional brute-force scanning with efficient targeted lookups. The library is partitioned into three domain-specific sub-libraries: Scene, Object, and Action, each utilizing normalized semantic concepts as primary keys.

In our implementation, each key points to a collection of video metadata entries. As shown in Figure~\ref{fig:library_content}, which visualizes a subset of the Scene library and Object library, each entry consists of a unique video ID and a concise caption generated by the Description Agent. This structure ensures that the textual granularity of the library is strictly aligned with potential user queries, facilitating high-precision matching while significantly reducing the number of candidates for the final fine-grained verification stage.

\begin{table}[t]
    \centering
    \small
    \setlength{\tabcolsep}{4pt} 
    \caption{Preliminary selection of the Description Agent.}
    \begin{tabular}{l|cc}
        \toprule
        \textbf{Model} & \textbf{Format Accuracy} & \textbf{Alignment Score} \\
        \midrule
        GPT-4.1 & 96.4\% & 83.1 \\
        Gemini-2.5-Pro & 97.8\% & 82.3 \\
        \textbf{Qwen3-omni-flash} & \textbf{98.5\%} & \textbf{84.6} \\
        \bottomrule
    \end{tabular}
    \label{tab:model_selection}
\end{table}


\section{Appendix: Prompt Settings}
\label{sec:appendix}

Since MAVIS operates as a training-free framework, the zero-shot reasoning and tool-invocation capabilities of the underlying MLLMs are primarily governed by precise prompt engineering. As summarized in Table~\ref{tab:agents}, we design specialized instruction sets for each agent to ensure a robust workflow. Rather than relying on parameter updates, MAVIS elicits different functional behaviors through role-specific prompts, enabling the planner to decompose user queries into semantic sub-intents, the description agent to produce concise structured video descriptions, and the scene, object, and action agents to retrieve and verify candidates from their corresponding semantic sub-libraries. 

These prompts are designed with two goals in mind. First, they encourage semantic consistency across different stages of the pipeline by enforcing standardized outputs and reducing irrelevant or overly verbose generations. Second, they improve the stability of multi-agent collaboration by clearly specifying each agent's scope, decision criteria, and expected response format, which is especially important for the subsequent veto-based debate process. We provide the main prompt templates in this appendix to improve reproducibility and to clarify how MAVIS translates general-purpose MLLMs into specialized retrieval agents without additional training.

\begin{table*}[t]
\centering
\small
\renewcommand{\arraystretch}{1.2} 
\caption{Agent roles and task specifications in MAVIS. Each agent has a specialized role: Description Agent parses videos into structured semantic components; Planner decomposes queries; Retrieval Agents independently nominate candidates; Debate Controller applies a veto protocol to filter logical mismatches.}
\label{tab:agents}
\begin{tabularx}{\textwidth}{l|X}
\toprule
\textbf{Agent / Role} & \textbf{Instruction and Task Specification} \\ \midrule

\textbf{Description Agent} & \textbf{System:} Act as a video description agent specialized in generating concise, query-aligned captions. The output should be concise and intent-focused. Avoid including excessive details, background information, or atmospheric nuances that are not central to the main action. \\
(Video Parsing) & \textbf{Task:} Analyze video frames to produce two outputs: (1) A \textit{Concise Caption} (5-20 words, matching the granularity of typical user queries, focusing on PRIMARY action, main objects, and key scene); (2) \textit{Structured Semantic Components} (Scene, Object, Action) normalized into canonical forms. Scene identifies primary location. Object extracts main entities in canonical form (singular, e.g., "dog", "cat", "person", "motorcycle"). Action extracts primary activity in gerund form (e.g., "playing", "riding", "cleaning"). Use standardized keywords and normalize synonyms. \\
& \textbf{Output Format:} JSON with \texttt{\{caption, scene, object, action\}} keys, where \texttt{scene}, \texttt{object}, and \texttt{action} are arrays of standardized keywords. \\ \midrule

\textbf{Planner} & \textbf{System:} Act as a query planner that decomposes user search queries into atomic semantic components. Identify which semantic dimensions (Scene, Object, Action) are present in the query and extract the corresponding sub-intents for each active dimension. \\
(Decomposition) & \textbf{Task:} (1) Identify which semantic dimensions are present: Scene, Object, and/or Action. (2) Extract the specific sub-intent for each active dimension. (3) If a dimension is not mentioned, exclude it from the output (selective activation). For example, "a dog running on the grass" activates Scene, Object, and Action; "a dog running" activates only Object and Action. \\
& \textbf{Output Format:} JSON containing \texttt{active\_dimensions} (list of activated roles, e.g., \texttt{["Object", "Action"]}) and \texttt{sub\_intents} (dictionary mapping each role to its sub-intent, e.g., \texttt{\{"Object": "dog", "Action": "running"\}}). \\ \midrule

\textbf{Retrieval Agent} & \textbf{System:} Act as a domain specialist (Scene/Object/Action) retrieval agent. Each agent focuses on one semantic dimension to search the semantic library for videos matching the query's component in that dimension. \\
(Candidate Proposal) & \textbf{Task:} (1) \textit{Category Matching:} Navigate the library index to identify the canonical category key that semantically matches the query sub-intent, resolving linguistic variations between query and stored canonical keys (e.g., "bike" → "motorcycle", "chatting" → "talking"). If no exact match, find the closest semantic category. (2) \textit{Candidate Retrieval:} Retrieve ALL videos from the matched category. (3) \textit{Similarity Computation:} For each candidate video, invoke a similarity tool to compute confidence scores $\phi_r(v_i) = \cos(E_{\text{VLM}}^{\text{txt}}(\text{query\_sub\_intent}), E_{\text{VLM}}^{\text{txt}}(\text{video\_caption}))$, where $E_{\text{VLM}}^{\text{txt}}(\cdot)$ is the VLM text encoder and $\cos(\cdot)$ computes cosine similarity. This quantifies the agent's confidence in the match. (4) \textit{Proposal Generation:} Generate a proposal message $M_r$ containing all videos where $\phi_r(v_i) > \tau_{\text{soft}}$ (threshold 0.5, designed to maximize recall). Act as a strict evaluator, computing confidence scores for each video. \\
& \textbf{Output Format:} JSON with \texttt{matched\_category} (the canonical category key), \texttt{reasoning} (explanation of why this category matches the query intent), and \texttt{proposal} (list of \texttt{\{id, cap, similarity\}} tuples for all candidates that pass the soft threshold). \\ \midrule

\textbf{Debate Controller} & \textbf{System:} Act as a logic-aware debate coordinator. Facilitate collaborative filtering among multiple retrieval agents using a strict \textit{Veto Protocol} to filter out logically inconsistent candidates. \\
(Veto Mechanism) & \textbf{Task:} (1) \textit{Complete Information:} Ensure every agent has a "vote" (score) on every candidate in the unified pool $\mathcal{V}_{\text{pool}}$. If an agent hasn't computed a score for a candidate, it performs an on-demand check. (2) \textit{Veto Protocol:} Apply logical negation: a video is VETOED (removed) if ANY agent identifies a hard conflict, where $\phi_r(v_i) < \tau_{\text{hard}}$ (threshold 0.3). (3) \textit{Controversial Candidates:} Identify videos that survive the veto. These are semantically plausible but require fine-grained verification. They have no hard conflicts from any agent, but may have mixed confidence scores. \\
& \textbf{Output Format:} JSON with \texttt{vetoed\_candidates}, \texttt{controversial\_candidates}, \texttt{veto\_reasons} and \texttt{controversial\_scores}. \\ 
\bottomrule
\end{tabularx}
\end{table*}

\end{document}